\pdfoutput=1

\documentclass[11pt]{article}

\usepackage[final]{acl}

\usepackage{times}
\usepackage{latexsym}
\usepackage{subfigure}

\usepackage[T1]{fontenc}

\usepackage[utf8]{inputenc}

\usepackage{microtype}

\usepackage{inconsolata}

\usepackage{graphicx}

\usepackage{xspace}
\newcommand{\wait}{\textit{wait}\xspace}
\newcommand{\mypar}[1]{\textbf{#1.}~}
\usepackage{amsmath}
\usepackage{amssymb}
\usepackage{mathtools}
\usepackage{amsthm}
\usepackage[capitalize,noabbrev]{cleveref}
\theoremstyle{plain}

\theoremstyle{definition}

\theoremstyle{remark}

\usepackage[textsize=tiny]{todonotes}

\title{Internal states before \wait modulate reasoning patterns}

\author{
 \textbf{Dmitrii Troitskii *\textsuperscript{1,2}},
 \textbf{Koyena Pal *\textsuperscript{2}},
 \textbf{Chris Wendler\textsuperscript{2}},
 \textbf{Callum Stuart McDougall},
 \textbf{Neel Nanda}
\\
 \textsuperscript{1}Independent,
 \textsuperscript{2}Northeastern University
\\
 \small{
   \textbf{Correspondence:} \href{mailto:email@domain}{troitskii.d@northeastern.edu, pal.k@northeastern.edu}
 }
}

\begin{document}
\maketitle
\begin{abstract}
Prior work has shown that a significant driver of performance in reasoning models is their ability to reason and self-correct.
A distinctive marker in these reasoning traces is the token \wait, which often signals reasoning behavior such as backtracking. 
Despite being such a complex behavior, little is understood of exactly why models do or do not decide to reason in this particular manner, which limits our understanding of what makes a reasoning model so effective. In this work, we address the question whether model's latents preceding \wait tokens contain relevant information for modulating the subsequent reasoning process. We train crosscoders at multiple layers of \texttt{DeepSeek-R1-Distill-Llama-8B} and its base version, and introduce a latent attribution technique in the crosscoder setting. We locate a small set of features relevant for promoting/suppressing \wait tokens' probabilities. Finally, through a targeted series of experiments analyzing max-activating examples and causal interventions, we show that many of our identified features indeed are relevant for the reasoning process and give rise to different types of reasoning patterns such as \emph{restarting from the beginning, recalling prior knowledge, expressing uncertainty, and double-checking}.



\end{abstract}
\section{Introduction}\label{introduction}

A growing class of language models known as reasoning models, for instance, DeepSeek-R1~\cite{deepseekai2025deepseekr1incentivizingreasoningcapability}, OpenAI o1 series~\cite{openai2024openaio1card}, and others~\cite{qwq-32b-preview, gemini-2.5, claude-3.7}, produce detailed internal reasoning chains before generating responses. While they showcase sophisticated reasoning capabilities, our understanding of their internal reasoning mechanisms remains limited. A recent work by~\citet{venhoff2025understanding} categorized DeepSeek R1's reasoning process into behavioral patterns such as example testing, uncertainty estimation, and backtracking to show how models solve reasoning tasks. When and how do these reasoning patterns form? 

This is a broad question that encompasses both the training dynamics, such as when and how reasoning circuits form, and the emergence of reasoning patterns and behaviors during inference. In this work, we focus on the latter, specifically within the distilled R1 model \texttt{DeepSeek-R1-Distill-Llama-8B}.\footnote{Although we work with the distilled model, we will refer to it as R1 for succinctness.} 
From observation, in this model the token \wait is strongly associated with the self-reflection of the model~\cite{baek2025towards}, which could relate to reasoning patterns such as backtracking, deduction, and uncertainty estimation~\cite{venhoff2025understanding}. Hence, we hypothesize that the features that modulate this token prediction can be useful in understanding the shifts in the reasoning patterns of the reasoning model. 
\begin{figure}[t]
    \centering
    \includegraphics[width=\linewidth]{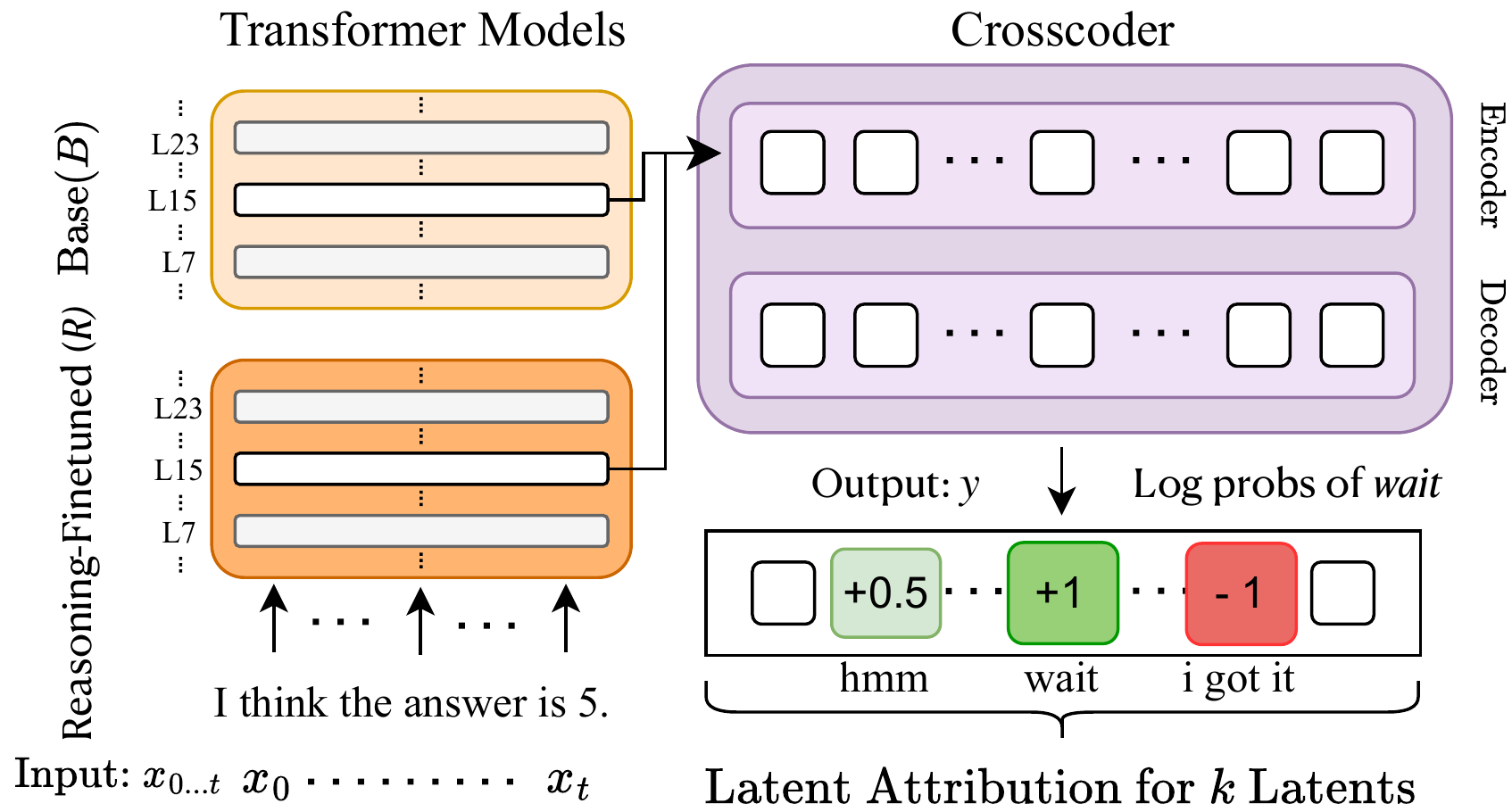}
    \caption{\textbf{We determine reasoning related crosscoder features} via latent attribution with respect to \wait tokens' logits.}
    \label{fig:attr_patching}
\end{figure} 

\mypar{Contributions} Motivated by this observation, we train sparse crosscoders~\citep{crosscoder2024} a recent mechanistic interpretability technique that allows to learn features within a paired base and finetuned model in an unsupervised way and can be used to perform model diffing. 

We leverage our crosscoders to discover latent features that modulate reasoning behaviors. To reduce the search space, our case study focuses on features promoting and suppressing \wait tokens that we hypothesize to play a key role in R1's reasoning. To this end, we extend attribution patching~\citep{attribution-patching} to the crosscoder setting. Our crosscoder latent attribution method allows us to efficiently score all of our 32,768 features regarding their contribution to the logits of \wait tokens. As a result, we obtain two short lists of relevant features, the 50 top features with the largest contributions to \wait logits and the 50 bottom ones, that we can effectively investigate.

By utilizing the crosscoders classification of the features in to \texttt{base, shared}, and \texttt{reasoning-finetuned} we found that surprisingly many reasoning related features fall inside of the bottom bucket, suggesting that both amplifying as well as suppressing \wait are of key importance to reasoning. By studying max-activating examples we observe a similar pattern. While top features mostly occur in the frequently reported backtracking and self-verification behaviors, we again notice that bottom features stand out in terms of the diversity of the reasoning behaviors within which they occur (e.g., starting again from an earlier step, expressing uncertainty, wrapping up, etc.). Finally, we perform a causal analysis via activation steering, in which we observe effects matching our interpretations based on the max-activating examples for many features, in particular the ones mentioned in parenthesis before.
Our work showcases how to reduce the combinatorial complexities of crosscoder feature-based interpretability with the help of our novel crosscoder latent feature attribution technique. We thereby open the door for future researchers to explore such findings in other chat and reasoning models.
\section{Methods}\label{methods}
To discover features, we create reasoning data instances (\cref{finding-reasoning-data}) and train sparse crosscoders (\cref{crosscoder-info}). We then apply latent attribution patching (\cref{latent-attr-patching}) to filter features that strongly modulate \wait tokens. Finally, we steer these features (\cref{steering}) to evaluate their impact on the model's reasoning behavior, helping us determine how actionable they are.

\subsection{Sparse Crosscoders}
\label{crosscoder-info}

Sparse autoencoders (SAEs) decompose model activations into a sparse set of linear feature directions. sparse crosscoders~\cite{crosscoder2024} extend this idea by jointly decomposing activations drawn from different layers, models, or context positions. To examine feature correspondences between two models, we train $\ell_1$-regularized sparse crosscoders between \texttt{DeepSeek-R1-Distill-Llama-8B} and its base model at the residual stream after layer 15. 

For an input sequence $x = (x_1, \dots, x_t)$ and token position $j \leq t$, we denote the residual stream using $\mathbf{a}(x_{\leq j}) \in \mathbb{R}^{d_{model}}$. We use $x_{\leq j} = (x_1, \dots, x_j)$ as an argument to highlight that the $j$th token's representation depends on all previous ones because of the attention layers within the model. To specify whether we use the base or the reasoning model we use the superscripts $\cdot^{(A)}$ and $\cdot^{(B)}$.

\mypar{Encoder} The Crosscoder activation at token $j$ is computed as
\begin{align}
    \mathbf{f}(x_{\leq j}) = \text{ReLU}\Big(\sum_{i \in \{B,R\}} W^{(i)}_{\text{enc}}\mathbf{a}^{(i)}(x_{\leq j}) + \mathbf{b}^{(i)}_{\text{enc}}\Big),
\end{align}
where $W_{\text{enc}}^{(i)} \in \mathbb{R}^{d_{\text{crosscoder}} \times d_{\text{model}}}$ and $\mathbf{b}_{\text{enc}}^{(i)} \in \mathbb{R}^{d_{\text{crosscoder}}}$ denote the trainable encoder parameters.  

\mypar{Decoder} The corresponding reconstruction for each model is given by
\begin{align}
    \mathbf{a}'^{(i)}(\mathbf{f}(x_{\leq j})) = W^{(i)}_{\text{dec}} \mathbf{f}(x_{\leq j}) + \mathbf{b}^{(i)}_{\text{dec}},
\end{align}
where $W^{(i)}_{\text{dec}} \in \mathbb{R}^{d_{\text{model}} \times d_{\text{crosscoder}}}$ and $\mathbf{b}_{\text{dec}}^{(i)} \in \mathbb{R}^{d_{\text{model}}}$ are the corresponding decoder parameters.  

For training details see Appendix~\ref{app:training}

\subsection{Data: Reasoning Instances with \wait}\label{finding-reasoning-data}
~\citet{venhoff2025understanding} released rollouts on reasoning problems from \texttt{DeepSeek-R1-Distill-Llama-8B} that have reasoning traces across ten categories that include topics like mathematical logic, creative problem solving, and scientific reasoning. 193 out of 500 samples have either of the following \textit{wait} tokens --- ``Wait", `` Wait", `` wait", ``wait". For every \textit{wait} in every sample that has a preceding reasoning sequence, we create a subsequence from the beginning of the sentence to the token right before \textit{wait}. This filters for \textit{wait} tokens related to reasoning behavior. This generates 350 subsequences.

\subsection{Crosscoder Latent Attribution}\label{latent-attr-patching}

We estimate the contribution of crosscoder feature to the metric 
\begin{align}\label{eq:mpatch}
    M(x) &= \sum_{y \in Y_\textit{wait}}p_{\theta}(y | x),
\end{align}
where $p_{\theta}(y | x)$ denotes the next token distribution defined by the reasoning-finetuned model, $x$ is a rollout that contains at least one \wait token and is truncated immediately before its occurrence and $Y_{wait}$ the set containing all \wait tokens, including, e.g., ``Wait",`` Wait", `` wait", ``wait".

In order to do so, we perform a zero ablation for each component $f_j$ at the last token
\begin{align}\label{eq:appendix_attr_patching}
    m_j(x) = M(x) - M(x|\text{do}(f_j \leftarrow 0)),
\end{align}
where $M(x|\text{do}(f_j \leftarrow 0))$ denotes the metric when the $j$th feature activation is set to zero. If $m_j$ is positive, ablating the feature decreases the probability of \wait. Thus, $m_j$ can be thought of as the $f_j$-s contribution to the probability of \wait. 

Since performing this zero ablation for each feature individually is computationally expensive and would require~$d_{crosscoder} + 1$ forward passes per datapoint $x$, we leverage linear approximations~\citep{Nanda2023AttributionPatching, marks2025sparse} to compute them efficiently. In particular, we use 
\begin{equation}\label{eq:attribution}
    \widehat{\mathbf{m}}(x) = {W_{dec}^R}^T \nabla_{\mathbf{a}} M(x) \odot\mathbf{f}(x),
\end{equation}
where $W_{dec}^R \in \mathbb{R}^{d_{model} \times d_{crosscoder}}$ are the parameters of the decoder, $\nabla_{\mathbf{a}} M(x) \in \mathbb{R}^{d_{model}}$ is the gradient of our metric on $x$'s last token and $\mathbf{f}(x)$ the crosscoder latents. The gradient is taken with respect to the elements of the residual stream $\mathbf{a}(x)$. We have $m_j(x) \approx \widehat{m}_j(x)$. 

We average our scores over a dataset of 620 rollouts ending right before the first \wait occurrence.

\subsection{Steering}\label{steering}

After training, the columns of the decoder matrix  $W^{(R)}_{dec} \in \mathbb{R}^{d_{model} \times d_{crosscoder}}$ can be thought of as $d_{crosscoder}$ linear feature directions/features $W_{dec}^{(R)} = (\mathbf{v}_1, \dots, \mathbf{v}_{d_{crosscoder}})$,
where $R$ denotes the reasoning model, $d$ is the dimension of the activations being decomposed, and  $d_{crosscoder}$ is the number of features extracted by our crosscoder.
These features can then be used to modulate the model's rollouts by adding them to the model's activations ($\mathbf{a}(R)$) during token generation:
\begin{equation}
\small
\mathbf{a}^{(R)}_{\mbox{steered}}(x)_i = \begin{cases}
    \mathbf{a}^{(R)}(x)_i + \alpha \frac{\|\mathbf{a}^{(R)}(x)\|}{\|\mathbf{v}_k\|}\mathbf{v}_{k},& \text{if i $\geq$ t}\\
    \mathbf{a}^{(R)}(x)_i, & \text{else}
\end{cases}
\end{equation}

in which we ``steered'' with strength $\alpha \in \mathbb{R}$ and the $k$th feature at the last token of the input $x_1, \dots, x_t$ and each newly generated token position $x_{t+1}, \dots$. We steer at layer 15.


\section{Results}\label{results}



\begin{figure}[!htb]
    \centering
    \includegraphics[width=\linewidth]{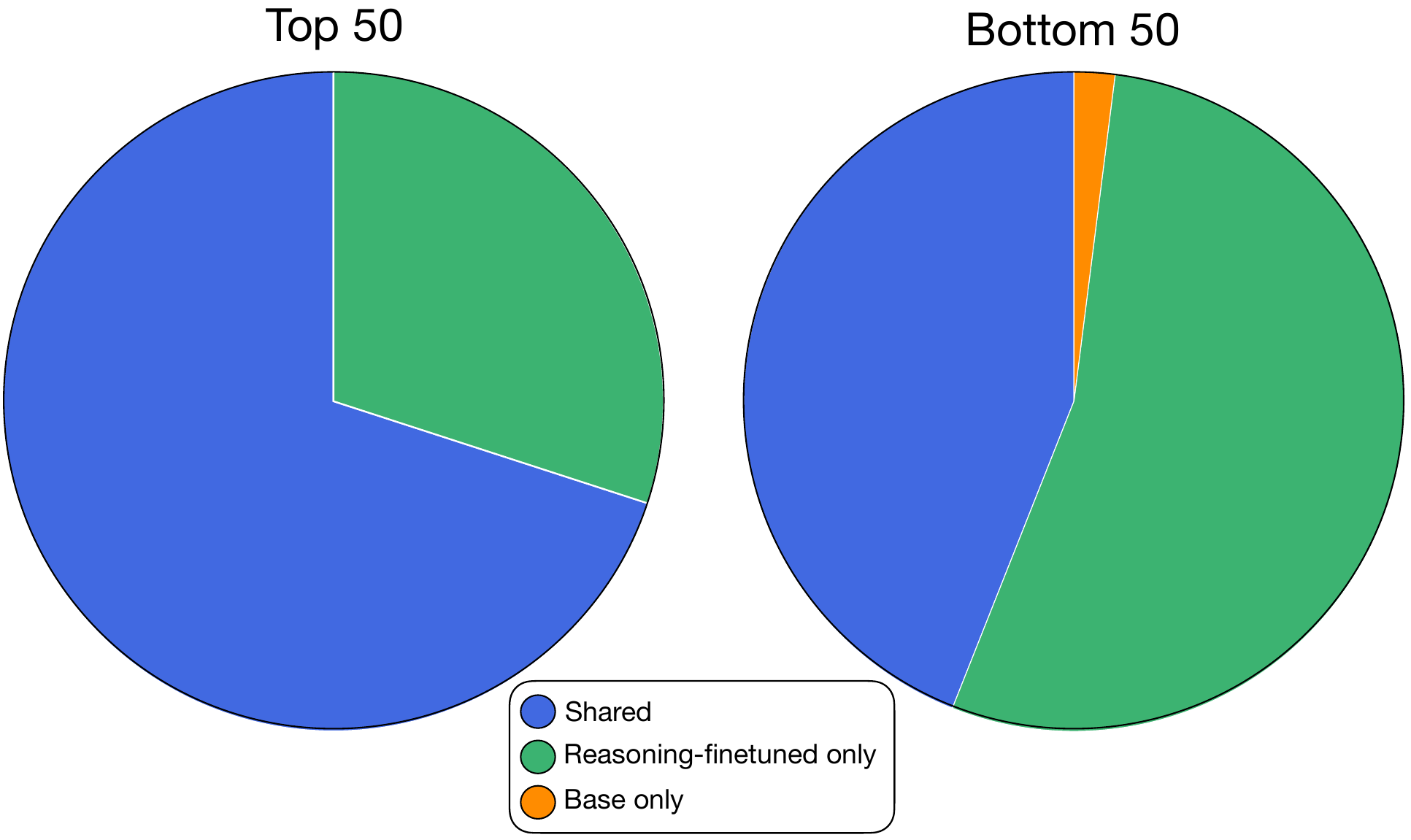}
    \caption{\textbf{Crosscoder classification of the features into \texttt{base, shared, reasoning-finetuned} categories}, with distribution attributed using the crosscoder relative norm difference}
    \label{fig:base-shared-ft}
\end{figure}

\mypar{Model diffing} Crosscoders by design allow to classify their learned features into \texttt{base}, \texttt{shared}, and \texttt{finetuned}. As \citet{minder2025latent} have recently shown this classification is not perfect, we ask the reader to take ~\cref{fig:base-shared-ft} with a grain of salt.~\cref{fig:base-shared-ft} shows the fraction of top and bottom features classified into the three categories. Both top and bottom features contain a significant number of \texttt{shared} features, which is expected, since most of the features learned by the crosscoder are in this category. The bottom features also contain some \texttt{base}-only features, - we hypothesize that some of the features that only the base model uses would decrease the likelihood of the \wait token in reasoning sequences, since the base model was not trained to predict the \wait token in such context. 

As expected, none of the top features are \texttt{base}-only. Most interestingly, the bottom features contain a larger number for \texttt{finetuned}-only, i.e., \texttt{reasoning} features. This suggests reasoning tuning allocates a substantial number of features to both suppressing as well as promoting \wait token's probabilities.

\mypar{Max activating examples} Next, we examine the features' max activating examples, which we obtain by computing feature coefficients over a dataset of 20 million tokens. Max-activating examples are input samples of various lengths from the crosscoder training data that elicit the highest activation of a given feature. They are identified by running the trained crosscoder on 200,000 validation samples, recording each feature’s activation, and selecting the top 100. For each sample, we highlight the most activating tokens, aiding interpretation. Top features most positively influence the logprob of the \wait token, while Bottom features have the strongest negative effect. We manually expect a set of 100 max activating examples and generate an automated annotation for each of our 100 features.

We observe that the Top features largely contribute to backtracking (max activating examples activate on \wait and "But" tokens). Bottom features correspond to behavior which can be interpreted as the model restarting its thinking process or concluding the reasoning trace. For example, feature \#1565 activates on full stops at the end of the reasoning sequence and feature \#32252 activates on a final answer at the end of samples with mathematical reasoning.

\begin{figure}
\includegraphics[width=1.\linewidth]{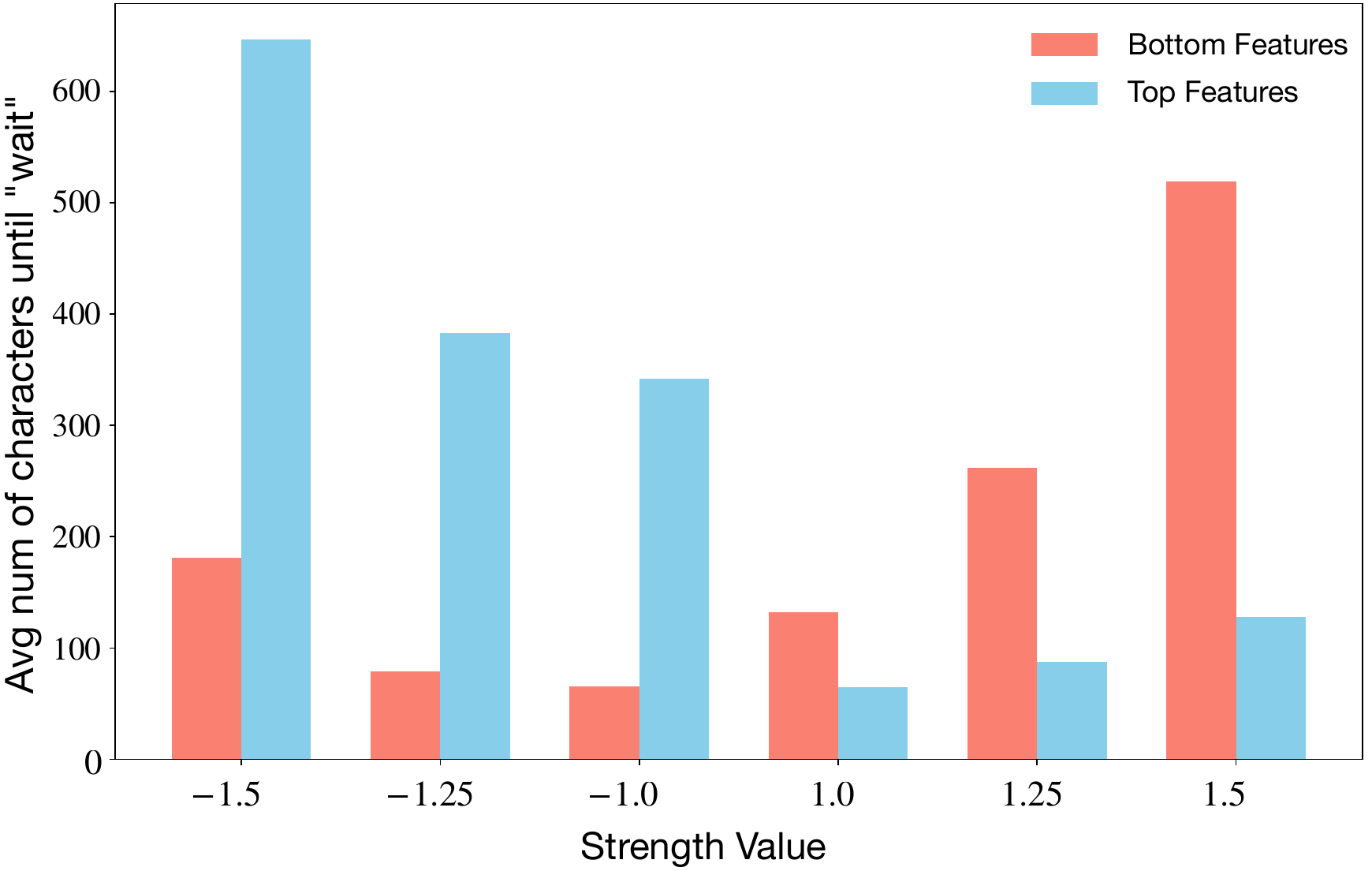}
\caption{\textbf{We compute how many characters occur before the first \wait token in the continued rollout}, while steering with all of our features and different intervention strengths. Steering with positive coefficient for the top features slightly increases the distance to the first \wait which is due to oversteering, for bottom features as one would expect the distance to the first \wait increases significantly. Negative steering has the opposite effect.}\label{fig:steering}
\end{figure} 

\begin{figure*}[!htb]
    \centering
    \includegraphics[width=1.\linewidth]{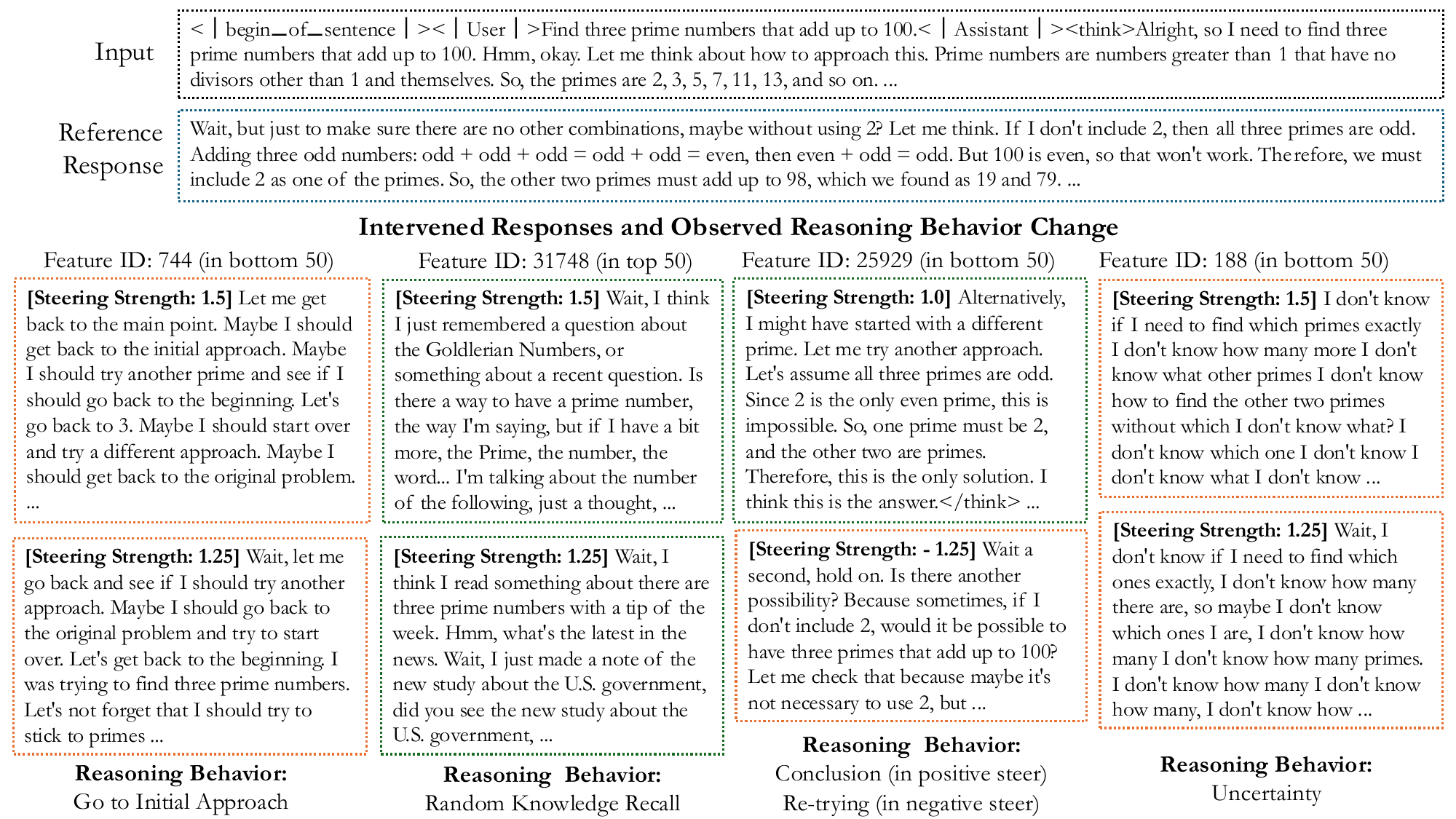}
    \caption{Change in \textbf{reasoning behavior observed when features are steered} in the positive and/or negative directions.}
    \label{fig:intervened-response}
\end{figure*}

\mypar{Feature steering} Finally, we investigate features' causal impact on the rollouts in the reasoning model by performing steering. This is done by activating a feature,  represented as a column vector of our learned decoder matrix, and multiplying it with a steering coefficient that sets its strength before adding it to the residual stream.
We steer on a single input example, see ~\cref{fig:intervened-response}. To match the setting in which we extracted the features we start steering the rollout from the token before the first \wait token and from there generate up to 200 tokens while steering. Since steering is sensitive to the steering strength hyperparameter, we steer with multiple strengths -1.5, -1.25, -1.0, -0.75, -0.5, 0.5, 0.75, 1, 1.25, 1.5 for each feature. This results in 10 continuations for each of the 100 features. 

In ~\cref{fig:steering}, we verified that the causal role of our features is broadly aligned with our expectations about them based on our feature selection criterion. In particular, we measure the number of characters occurring before the first \textit{wait}. Since we selected our features based on latent attribution with respect to \textit{wait}, for top features the number of characters should be small when steering positively and for bottom features it should be large. Similarly, for features that have meaningful effects in both directions—which is not guaranteed, since feature coefficients are always positive—steering with negative strength should increase the number of characters before \wait for top features and decrease it for bottom features. 

\mypar{Steered generations} 
We show example continuations under our interventions in   ~\cref{fig:intervened-response}. As can be seen, in particular, the bottom features lead to interesting reasoning patterns that we have not seen before in the literature. Among the top features we observed many instances of steering positively quickly resulting in degenerate sequences like ``WaitWaitWait...'' or `` wait wait wait...''. Steering these into the negative direction leads to \wait disappearing from the outputs. Our best guess is that those features literally contribute to the \wait tokens and it is unclear whether they can also trigger any specific reasoning patterns that would be comparable to the ones reported in ~\cref{fig:intervened-response}. Additionally, we found several features that when steered positively/negatively lead towards the model wrapping up and providing the final response and when steered into the opposite direction lead to extended reasoning.

\mypar{Downstream evaluation on \textsc{Math500}}  We compute rollouts for each sample with temperature=0.6, top\_p=0.95, and max\_tokens=7500, resulting in 86\% accuracy. Next, we filter for completions containing the \wait token. From these, we sample 100 problems with 81 correct / 19 incorrect rollouts. We steer with one feature at a time (IDs 188, 744, 25929, 31748) at layer 15 with strength \(\alpha = 1.5\), applied to the first 100 tokens (full-sequence steering degraded quality).

We evaluate using exact-match accuracy, median change in completion length, and LLM-judge adherence score (GPT-5-nano~\cite{openai_gpt5nano_model_doc_2025}). Accuracy remained close to 81\% baseline for three features but decreased to 61\% for 31748. Detailed results are provided in ~\cref{tab:math500_steer}, and judge implementation details are included in ~\cref{sec:prompt-appendix}. Manual inspection suggests the judge systematically underestimates adherence, and thus the reported scores should be interpreted as lower bounds.

\begin{table}[t]
\centering
\footnotesize
\setlength{\tabcolsep}{4pt}
\caption{\textbf{Math500 ($n{=}100$) downstream evaluation with feature steering} at layer 15 ($\alpha{=}1.5$) over first 100 tokens; Acc = exact-match accuracy; $\Delta$len = median token increase; Adh.\ is LLM-judge score}
\begin{tabular}{l c c r r r}
\hline
\textbf{Feature} & \textbf{Type} & \textbf{Acc} & \textbf{$\Delta$Length (med, tok) [\%]} & \textbf{Adh.} \\
\hline
188   & Bottom & 84\%  & +429 (+28\%) & 83 \\
744   & Bottom & 81\%  & +473 (+39\%) & 58 \\
25929 & Bottom & 79\%  & +310 (+21\%) & 31$^{\dagger}$ \\
31748 & Top    & 61\%  & +526 (+45\%) & 36 \\
\hline
\end{tabular}
\label{tab:math500_steer}
\end{table}

\section{Related Work}\label{related-work}

To understand reasoning models, ~\citet{venhoff2025understanding} looks into the chain-of-thought texts and categorizes the trace into various reasoning behavior patterns such as deduction, adding knowledge, and example testing. They further find steering vectors to increase or decrease the appearance of a particular reasoning behavior. ~\citet{baek2025towards} investigates reasoning features, learned by crosscoders, via qualitative analysis of max activating examples. An additional perspective on token-level and feature-level analysis of reasoning features is a concurrent work by ~\citet{lee2025geometryselfverificationtaskspecificreasoning} that performs a case study on the self-verification process of a task-specific reasoning model. 

Similarly,  \citet{zhang2025reasoning} show that reasoning models ``know'' when they are right by training linear probes that can predict the correctness of intermediate solutions from the models' internals with high accuracy. In contrast to \citep{venhoff2025understanding,lee2025geometryselfverificationtaskspecificreasoning,zhang2025reasoning} that focus on specific features and reasoning behaviors on labeled datasets, we are using crosscoders to discover features relevant to distilled reasoning model's ``thought'' process with minimal supervision. Compared \citet{baek2025towards} we focus on a more narrow set of the features that we obtain by our latent attribution technique that selects features most relevant for modulating the logits of different ``wait'' tokens.
\section{Conclusion}\label{discussion}
Our study sheds light on how the model's latents preceding \wait tokens signal behaviors like backtracking. We introduce latent attribution patching for identifying and testing which internal features influence these tokens in crosscoder models. We show that these features not only predict \textit{wait} tokens but also shape how the model reasons. 
\clearpage
\section*{Limitations}

While our method identifies and manipulates features influencing \textit{wait} tokens, it is currently limited to a specific model family and may not generalize across architectures without adaptation. Additionally, our analysis focuses on a narrow slice of reasoning behavior, potentially missing out other important markers or mechanisms. Finally, while we demonstrate causal influence on reasoning behavior, fully constructing a circuit that also shows latent-to-latent dependency across layers for these reasoning behaviors remains an open area for future work.

\section*{Data and Code Availability}
Crosscoder weights: \url{https://huggingface.co/mitroitskii/Crosscoder-Llama-3.1-8B-vs-Llama-R1-Distill-8B}. \\ Crosscoder training: \url{https://github.com/science-of-finetuning/crosscoder_learning}. \\ Crosscoders analysis and max activating examples: \\ \url{https://github.com/science-of-finetuning/sparsity-artifacts-crosscoders}. \\ Attribution experiments: \url{https://github.com/mitroitskii/interp-experiments/tree/main/reasoning_circuits}. \\ Steering experiments: \url{https://github.com/wendlerc/r1helpers}.

\section*{Acknowledgments}
This project began as part of the training phase of Neel Nanda’s MATS 8.0 stream. 
We are grateful to 
Neel and Arthur Conmy for their valuable feedback and suggestions during that time. We would also like to thank Clément Dumas and Julian Minder for their helpful guidance on crosscoder setup, Caden Juang for advice on the attribution setup, Constantin Venhoff for providing the dataset of prompts, Andy Arditi for his valuable feedback as well as the Bau Lab for providing compute resources.


\bibliography{main}
\appendix
\onecolumn
\section{Notations}
In this section, we summarize the different notations we use for our mathematical expressions and equations. Usually, we refer to vectors using bold letters and other scalar components, including vector components, as regular (non-bold) characters.
\begin{enumerate}
    \item $x = (x_0, ..., x_i)$ is input to the base and reasoning-finetuned models
    \item $\mathbf{a}$ is the residual stream of the model
    \item $d_{model}$ is dimensionality of the residual stream and $\mathbf{a} \in \mathbb{R}^{d_{model}}$
    \item $\cdot^{(R)}$ and $\cdot^{(B)}$ refers to reasoning-finetuned and base model components.
    \item $\mathbf{f}$ refers to the feature coefficients or feature activations of the crosscoder. In other words, it refers to the latents of the crosscoder.
    \item $W_{enc}$ denotes encoder matrix; $W_{dec}$ the decoder matrix, and $b_{enc}, b_{dec}$ the corresponding biases
    \item $\mathbf{v}_j$ refers to the features learned by the crosscoder.
    \item $d_{crosscoder}$ denotes the number of features learned by the crosscoder.
    \item $M$ is the scalar metric that measures how much a model promotes or suppresses \wait
\end{enumerate}

\section{Crosscoder Training}\label{app:training}

The Crosscoder is trained to minimize a loss comprising a reconstruction mean squared error (MSE) term and an $\ell_2$ penalty on the decoder weights, weighted by feature activations:
\begin{align}
    L = \sum_{i \in \{B,R\}} \|\mathbf{a'}^{(i)} - \mathbf{a}^{(i)}\|^{2} 
      + \sum_{k} f_{k}(x) \sum_{i \in \{B,R\}} \|W^{(i)}_{\text{dec},k}\|.
\end{align}

We train three Crosscoders of this form for 18 hours using 50–100\% of 8×A100 GPUs.

\section{Patchscope Experiments}

As an additional check we investigate our selected features through the patchscope-lens~\cite{ghandeharioun2024patchscopes}. It is an intermediate decoding technique that decodes (some) of the information contained in a latent by inserting it at its corresponding layer in a parallel forward pass that is processing a patchscope-lens prompt. Since we are using a reasoning model, we slightly adapt the patchscope-lens prompt to the reasoning model's format and thought prefilling: 

\label{sec:patchscope-appendix}
\begin{verbatim}
<begin of sentence><User>Continue the 
following pattern: cat cat
1135 1135
hello hello
<Assistant><think>
Okay I need to complete: cat cat
1135 1135
hello hello
?
\end{verbatim}

\begin{figure} [!htb]
\centering
\includegraphics[width=0.4\textwidth]{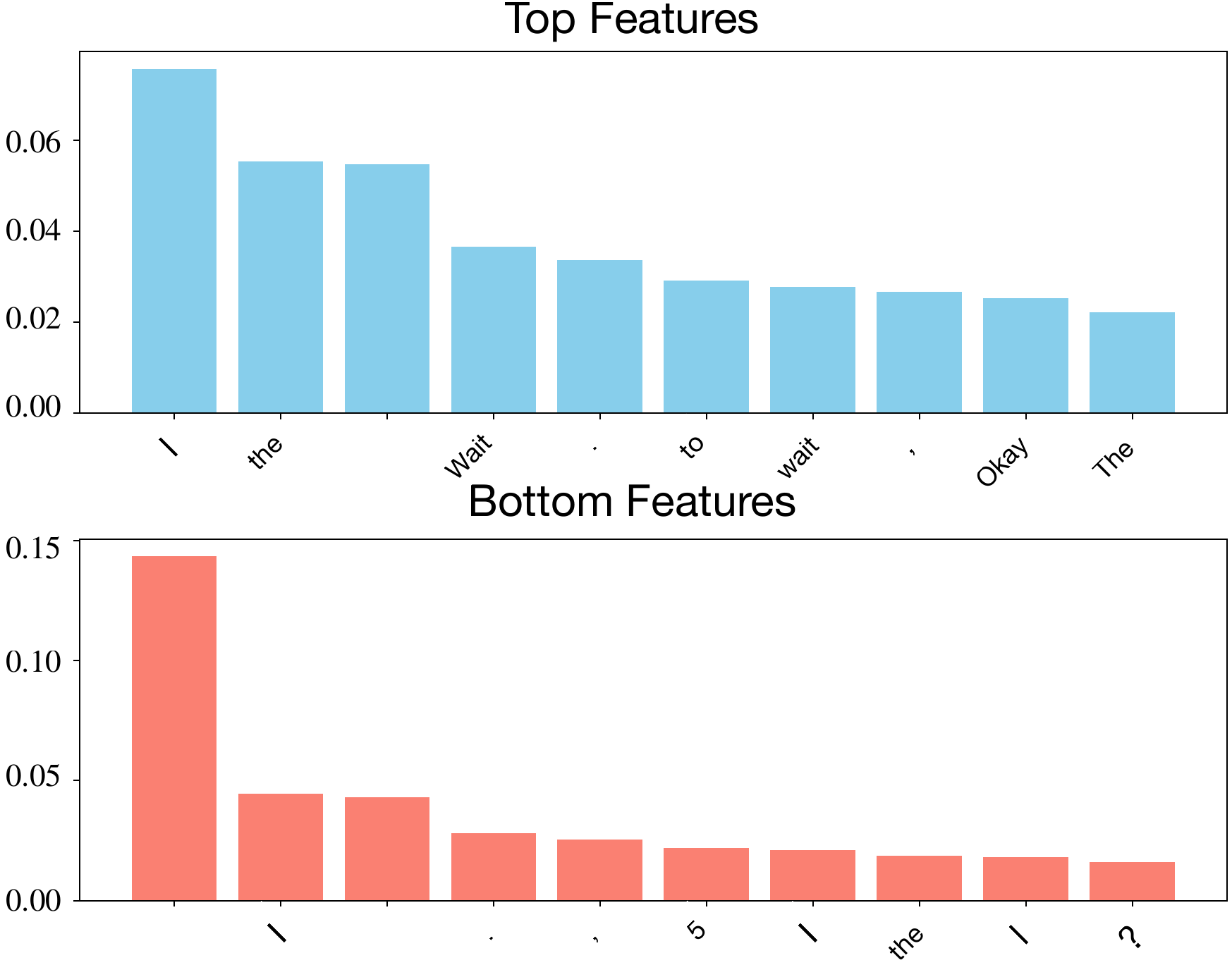}
\caption{\textbf{Top token probabilities from patchscope:} top features promote 'Wait' and related tokens, while bottom features show no clear pattern.}\label{fig:patchscope}
\end{figure}

For each of the top/bottom features, we compute the patchscope's next token distribution and take the average within the top/bottom groups. As a result, we obtain ~\cref{fig:patchscope}. As can be seen top features indeed promote ``Wait'' and related tokens, whereas, bottom ones share `` '' as top token. 

\section{LLM Judge Prompt Template}\label{sec:prompt-appendix}

\begin{verbatim}
"You are an expert evaluator of language model behavior and steering interventions.

Please analyze the following example and assess how much the steered response 
corresponds to the steering direction "{steering_direction}" 
on a scale from 1 to 100, where:

- 1-20: Very little to no correspondence with the steering direction
- 21-40: Some correspondence but not strong
- 41-60: Moderate correspondence
- 61-80: Strong correspondence
- 81-100: Very strong/excellent correspondence with the steering direction

**Original Context:**
{text_before}

**Original Response (before steering):**
{original_text_after}

**Steered Response (after steering):**
{steered_text_after}

**Assessment Criteria:**
- Consider how well the steered response reflects the intended steering direction 
  "{steering_direction}"
- Look for behavioral changes, linguistic patterns, and content shifts
- Compare the steered response to the original response
- Evaluate the consistency and strength of the steering effect

Please provide your assessment as a JSON object with the following format:
{
    "score": <number between 1-100>,
    "reasoning": "<brief explanation of your assessment>"
}

Assessment:"
\end{verbatim}




\end{document}